%% file: main.tex
\documentclass[10pt,twocolumn,letterpaper]{article}

\input{notation}

\usepackage[pagenumbers]{cvpr} % To force page numbers, e.g. for an arXiv version

\input{preamble}

\usepackage[pagebackref,breaklinks,colorlinks,citecolor=cvprblue]{hyperref}

\def\paperID{5524} % *** Enter the Paper ID here
\def\confName{CVPR}
\def\confYear{2024}

\title{Harnessing Large Language Models for Training-free Video Anomaly Detection}

\author{
{Luca Zanella\textsuperscript{1}
\quad
Willi Menapace\textsuperscript{1}
\quad
Massimiliano Mancini\textsuperscript{1}
\quad
Yiming Wang\textsuperscript{2}
\quad
Elisa Ricci\textsuperscript{1,2}}\\[-4mm]\and
{University of Trento\textsuperscript{1}\quad Fondazione Bruno Kessler\textsuperscript{2}}\\[1mm]
{\tt\small \url{https://lucazanella.github.io/lavad/}}
}

\begin{document}
\maketitle
\input{sections/0_abstract}
\input{sections/1_intro}
\input{sections/2_related}
\input{sections/3_0_preliminary}
\input{sections/3_method}
\input{sections/4_experiment}
\input{sections/5_conclusion}

\vspace{10pt}
\noindent\textbf{Acknowledgments.}
This work is supported by MUR PNRR project FAIR - Future AI Research (PE00000013), funded by NextGeneration EU and by PRECRISIS, funded by EU Internal Security Fund (ISFP-2022-TFI-AG-PROTECT-02-101100539). We acknowledge the CINECA award under the ISCRA initiative, for the availability of high-performance computing resources and support.

\clearpage

{
    \small
    \bibliographystyle{ieeenat_fullname}
    \bibliography{main}
}

\clearpage
\setcounter{page}{1}
\appendix
\maketitlesupplementary

\input{supp/supp}
\input{supp/prompts}
\begin{figure*}[ht!]
\vspace{-2mm}
\centering
\includegraphics[width=\linewidth]{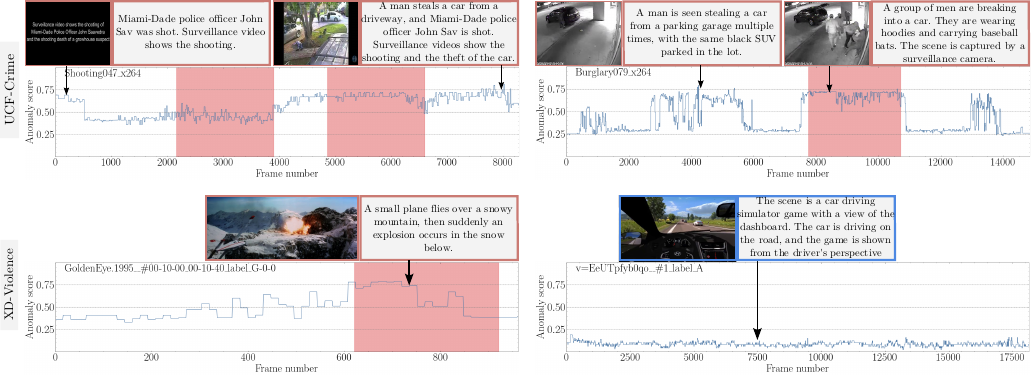}
  \caption{We showcase qualitative results obtained by \methodshort~on four test videos, including two videos (top row) from UCF-Crime and two videos from XD-Violence (bottom row). For each video, we plot the anomaly score over frames computed by our method. We display some keyframes alongside their most aligned temporal summary (blue bounding boxes for normal frame predictions and red bounding boxes for abnormal frame predictions), illustrating the relevance among the predicted anomaly score, visual content, and description. \inlineColorbox{AnomalyRed}{Ground-truth anomalies} are highlighted.}
 \label{fig:supp_qualitatives}
\end{figure*}
\input{supp/ablation}
\input{supp/qualitatives}
\input{supp/limitations}
\input{supp/broader_societal_impacts}

\end{document}

%% file: notation.tex
\newcommand{\classbinary}{y}

\newcommand{\suppmat}{Supp. Mat.}
\def\ie{\textit{i.e}\onedot} 

\newcommand{\netvlmtext}{\mathcal{E}_T}
\newcommand{\netvlmimage}{\mathcal{E}_I}
\newcommand{\netvlmvideo}{\mathcal{E}_V}

\newcommand{\netcaptioner}{\Phi_\mathtt{C}}
\newcommand{\netllm}{\Phi_\mathtt{LLM}}

\newcommand{\methodname}{\textbf{LA}nguage-based \textbf{VAD}}
\newcommand{\methodshort}{LAVAD}
\newcommand{\llama}{Llama}
\newcommand{\bliptwo}{BLIP-2}

\newcommand{\image}{\mathbf{I}}
\newcommand{\imagespace}{\mathcal{I}}
\newcommand{\videospace}{\mathcal{V}}
\newcommand{\languagespace}{\mathcal{T}}
\newcommand{\latentspace}{\mathcal{Z}}

\newcommand{\video}{\mathbf{V}}
\newcommand{\rawcaption}{C}
\newcommand{\rawcaptionset}{\mathbf{C}}
\newcommand{\rawcaptionsetclean}{\hat{\mathbf{C}}}
\newcommand{\retrievedcaption}{\hat{\rawcaption}}

\newcommand{\summarycaption}{S}

\newcommand{\prompt}{\texttt{P}}
\newcommand{\numimages}{M}

\newcommand{\numk}{K}

\newcommand{\trainingset}{\mathcal{D}}

\newcommand{\clipduration}{T}
\newcommand{\framesperclip}{N}

\newcommand{\scoreabn}{a}

\newcommand{\refinedscoreabn}{\tilde{a}}

\newcommand{\inlineColorbox}[2]{\begingroup\setlength{\fboxsep}{1pt}\colorbox{#1}{\hspace*{2pt}\vphantom{Ay}#2\hspace*{2pt}}\endgroup}

%% file: preamble.tex
\usepackage[accsupp]{axessibility} % Improves PDF readability for those with visual impairments.
\usepackage{booktabs}
\usepackage{multirow}
\usepackage{graphicx}
\usepackage{tikz}
\usepackage{amsmath}
\usepackage{colortbl}
\usepackage{xcolor}
\usepackage{pifont}
\usepackage{overpic}
\usepackage{microtype}

\definecolor{cvprblue}{rgb}{0.21,0.49,0.74}
\definecolor{VisionBlue}{RGB}{218,232,252}
\definecolor{LanguageOrange}{RGB}{255,230,204}
\definecolor{ModelGreen}{RGB}{213,232,212}
\definecolor{FunctionPurple}{RGB}{245,233,251}
\definecolor{AnomalyRed}{RGB}{255,170,170}
\definecolor{LightGreen}{RGB}{124,232,124}

\newcommand{\cmark}{\textcolor{LightGreen}{\ding{51}}}
\newcommand{\xmark}{\textcolor{red}{\ding{55}}}

%% file: sections/0_abstract.tex
\begin{abstract}
Video anomaly detection (VAD) aims to temporally locate abnormal events in a video. Existing works mostly rely on training deep models to learn the distribution of normality with either video-level supervision, one-class supervision, or in an unsupervised setting.  
Training-based methods are prone to be domain-specific, thus being costly for practical deployment as any domain change will involve data collection and model training.
In this paper, we radically depart from previous efforts and propose \methodname~(\methodshort), a method tackling VAD in a novel, \emph{training-free} paradigm, exploiting the capabilities of pre-trained large language models (LLMs) and existing vision-language models (VLMs). We leverage VLM-based captioning models to generate textual descriptions for each frame of any test video. With the textual scene description, we then devise a prompting mechanism to unlock the capability of LLMs in terms of temporal aggregation and anomaly score estimation, turning LLMs into an effective video anomaly detector.
We further leverage modality-aligned VLMs and propose effective techniques based on cross-modal similarity for cleaning noisy captions and refining the LLM-based anomaly scores. We evaluate \methodshort~on two large datasets featuring real-world surveillance scenarios (UCF-Crime and XD-Violence), showing that it outperforms both unsupervised and one-class methods without requiring any training or data collection.
% The code is publicly available at \url{https://github.com/lucazanella/lavad}.
\end{abstract}

%% file: sections/1_intro.tex
\section{Introduction} \label{sec:intro}
\begin{figure}[t!]
  \includegraphics[width=1\linewidth]{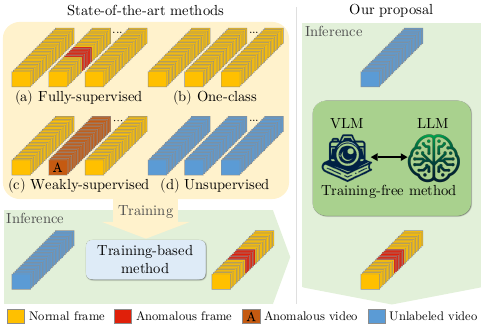}
  \caption{We introduce the first training-free method for video anomaly detection (VAD), diverging from state-of-the-art methods that are ALL training-based with different degrees of supervision. Our proposal, \methodshort, leverages modality-aligned vision-language models (VLMs) to query and enhance the anomaly scores generated by large language models (LLMs).}
 \label{fig:teaser}
\end{figure}

Video anomaly detection (VAD) aims to temporally localize events that deviate significantly from the normal pattern in a given video, \ie the anomalies. VAD is challenging as anomalies are often undefined and context-dependent, and they rarely occur in the real world.
The literature \cite{jiao2023survey} often casts VAD as an out-of-distribution detection problem and learns the normal distribution using training data with different levels of supervision (see \cref{fig:teaser}), including fully-supervised (\ie frame-level supervision of both normal and abnormal videos) \cite{bai2019traffic, wang2019anomaly}, weakly-supervised (\ie video-level supervision of both normal and abnormal videos)~\cite{sultani2018real, wu2021learning, tian2021weakly, li2022scale, li2022self, joo2023clip}, one-class (\ie only normal videos) \cite{park2020learning, liu2021hybrid, lv2021learning, sun2023hierarchical, yan2023feature,zaheer2020old}, and unsupervised (\ie unlabeled videos)~\cite{zaheer2022generative, tur2023exploring, tur2023unsupervised}. 
While more supervision leads to better results, the cost of manual annotation is prohibitive. On the other hand, unsupervised methods assume abnormal videos to constitute a certain portion of the training data, a fragile assumption in practice without human intervention.

Crucially, every existing method necessitates a training procedure to establish an accurate VAD system, and this entails some limitations. One primary concern is generalization: a VAD model trained on a specific dataset tends to underperform in videos recorded in different settings (e.g., \textit{daylight} versus \textit{night} scenes). Another aspect, particularly relevant to VAD, is the challenge of data collection, especially in certain application domains (\eg video surveillance) where privacy issues can hinder data acquisition. These considerations led us to explore a novel research question: \textit{Can we develop a training-free VAD method?}

In this paper, we aim to answer this challenging question. Developing a training-free VAD model is hard due to the lack of explicit visual priors on the target setting. However, such priors might be drawn using large foundation models, renowned for their generalization capability and wide knowledge encapsulation. Thus, we investigate the potential of combining existing vision-language models (VLMs) with large language models (LLMs) in addressing training-free VAD. 
On top of our preliminary findings, we propose the first training-free \textbf{LA}nguage-based \textbf{VAD} method (\textbf{\methodshort}), that jointly leverages pre-trained VLMs and LLMs for VAD. 
\methodshort~first exploits an off-the-shelf captioning model to generate a textual description for each video frame. We address potential noise in the captions by introducing a cleaning process based on the cross-modal similarity between captions and frames in the video.
To capture the dynamics of the scene, we use an LLM to summarize captions within a temporal window.
This summary is used to prompt the LLM to provide an anomaly score for each frame, which is further refined by aggregating the anomaly scores among frames with semantically similar temporal summaries.
We evaluate \methodshort~on two benchmark datasets: UCF-Crime \cite{sultani2018real} and XD-Violence \cite{wu2020not}, and empirically show that our training-free proposal outperforms unsupervised and one-class VAD methods on both datasets, demonstrating that it is possible to address VAD with \textit{no training and no data collection}. 

\vspace{5pt}
\noindent\textbf{Contributions.} In summary, our contributions are:
\begin{itemize}[noitemsep,nolistsep,leftmargin=*]
\item We investigate, for the first time, the problem of training-free VAD, advocating its importance for the deployment of VAD systems in real settings where data collection may not be possible. 

\item We propose \methodshort, the first language-based method for training-free VAD using LLMs to detect anomalies exclusively from a scene description. 

\item We introduce novel techniques based on cross-modal similarity with pre-trained VLMs to mitigate noisy captions and refine the LLM-based anomaly scoring, effectively improving the VAD performance.

\item Experiments show that, while using no task-specific supervision and no training, \methodshort~achieves competitive results w.r.t. unsupervised and one-class VAD methods, opening new perspectives for future VAD research.

\end{itemize}

%% file: sections/2_related.tex
\section{Related Work}
\label{sec:related_work}

\textbf{Video Anomaly Detection.} Existing literature on \textit{training-based} VAD methods can be categorized into four groups, depending on the level of supervision: supervised, weakly-supervised, one-class classification, and unsupervised. 
\textit{Supervised VAD} relies on frame-level labels to distinguish normal from abnormal frames~\cite{bai2019traffic, wang2019anomaly}. However, this scenario has received little attention due to its prohibitive annotation effort.
\textit{Weakly-supervised VAD} methods have access to video-level labels (the entire video is labeled as abnormal if at least one frame is abnormal, otherwise is regarded as normal)~\cite{sultani2018real, wu2021learning, tian2021weakly, li2022scale, li2022self, joo2023clip}. Most of these methods utilize 3D convolutional neural networks for feature learning and employ a multiple instance learning (MIL) loss for training. 
\textit{One-class VAD} methods train only on normal videos, although manual verification is necessary to ensure the normality of the collected data. Several methods \cite{park2020learning, liu2021hybrid, lv2021learning, sun2023hierarchical, yan2023feature,zaheer2020old} have been proposed, \eg considering generative models \cite{yan2023feature} or pseudo-supervised methods, where pseudo-anomalous instances are synthesized from normal training data \cite{zaheer2020old}. 
Finally, \textit{Unsupervised VAD} methods do not rely on predefined labels, leveraging both normal and abnormal videos with the assumption that most videos contain normal events
~\cite{zaheer2022generative, tur2023exploring, tur2023unsupervised, thakare2023dyannet, thakare2023rareanom}. 
Most methods in this category exploit generative models to capture normal data patterns in videos. In particular, generative cooperative learning (GCL) \cite{zaheer2022generative} employs alternating training: an autoencoder reconstructs input features, and pseudo-labels from reconstruction errors guide a discriminator. Tur \etal \cite{tur2023exploring, tur2023unsupervised} use a diffusion model to reconstruct the original data distribution from noisy features, calculating anomaly scores based on the reconstruction error between denoised and original samples. 
Other approaches \cite{thakare2023dyannet, thakare2023rareanom} train a regressor network from a set of pseudo-labels generated using OneClassSVM and iForest \cite{liu2012isolation}.

Instead, we completely sidestep the need for collecting data and training the model by exploiting existing large-scale foundation models to design a training-free pipeline for VAD.  

\vspace{5pt}
\noindent\textbf{LLMs for VAD}. Recently, LLMs have been explored in detecting visual anomalies across diverse application domains. Kim \etal \cite{kim2023unsupervised} propose an unsupervised method that mainly leverages VLMs for detecting anomalies, where ChatGPT is only utilized to produce textual descriptions that characterize normal and anomalous elements. However, the method involves human-in-the-loop to refine the LLM's outputs according to specific application contexts and requires further training to adapt the VLM. Other examples include exploiting LLMs for spatial anomaly detection in images addressing specific applications in robotics~\cite{elhafsi2023semantic} or industry~\cite{gu2023anomalygpt}.

Differently, we leverage LLMs together with VLMs to address temporal anomaly detection on videos and propose the first \textit{training-free} method for VAD, requiring no training and no data collection. 

%% file: sections/3_0_preliminary.tex
\section{Training-Free VAD} 
\label{sec:preliminary}
In this section, we first formalize the VAD problem and the proposed training-free setting (\cref{sec:pf}). We then analyze the capabilities of LLMs in scoring anomalies in video frames (\cref{sec:llms}). Finally, we describe \methodshort, our proposed VAD method (\cref{sec:method}).

\begin{figure}[t!]
  \includegraphics[width=1.0\linewidth]{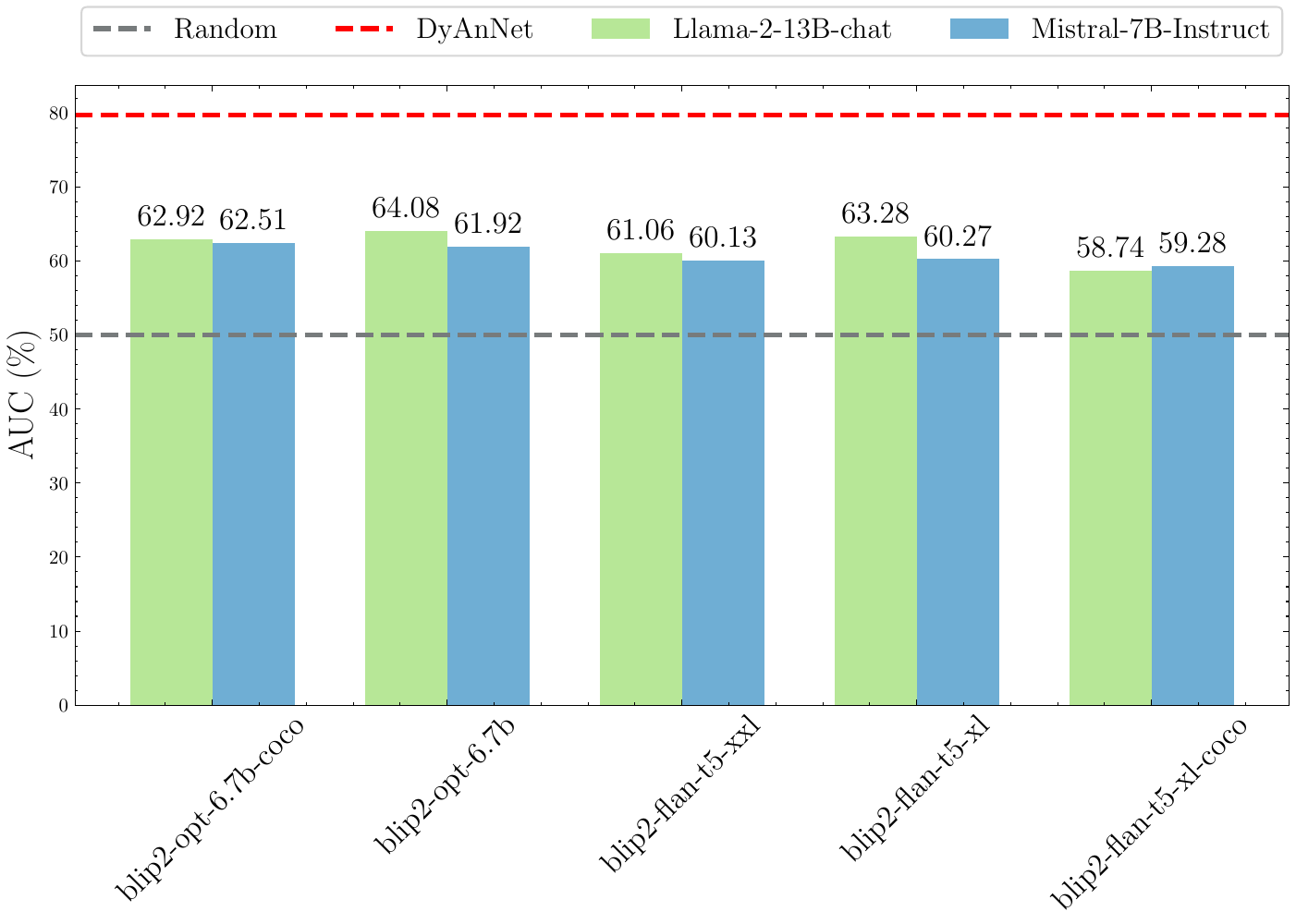}
  \caption{Bar plot of the VAD performance (AUC ROC) by querying LLMs with textual descriptions of video frames from various captioning models on the UCF-Crime test set. Different bars correspond to different variants of the captioning model \bliptwo~\cite{li2023blip}, while different colors indicate two different LLMs~\cite{touvron2023llama, jiang2023mistral}. For reference, we also plot the performance of the best-performing unsupervised method \cite{thakare2023dyannet} in a red dashed line, and that of a random classifier in a gray dashed line.}
 \label{fig:preliminary_bar}
\end{figure}

\begin{figure}[t!]
  \includegraphics[width=1.0\linewidth]{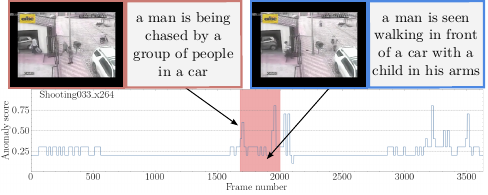}
  \caption{The anomaly score predicted by \llama~\cite{touvron2023llama} over time for video \textit{Shooting033} from UCF-Crime. We highlight some sample frames with their associated \bliptwo~captions to demonstrate that the caption can be semantically noisy or incorrect (red bounding boxes are for abnormal predictions while blue bounding boxes are for normal predictions). \inlineColorbox{AnomalyRed}{Ground-truth anomalies} are highlighted. In particular, the caption of the frame enclosed by a blue bounding box within the ground truth anomaly fails to accurately represent the visual content, leading to a wrong classification due to the low anomaly score given by the LLM.}
 \label{fig:preliminary_qualitative}
\end{figure}

\subsection{Problem formulation} 
\label{sec:pf}
Given a test video $\video = \left[\image_1, \ldots, \image_{\numimages}\right]$ of $\numimages$ frames, traditional VAD methods aim to learn a model $f$, which can classify each frame $\image \in \video$ as either normal (score 0) or anomalous (score 1), \ie $f:{\imagespace}^\numimages\rightarrow \left[ 0, 1 \right]^\numimages$ with $\imagespace$ being the image space. $f$ is usually trained on a dataset $\trainingset$ that consists of tuples in the form $(\video,~\classbinary)$. Depending on the supervision level, $\classbinary$ can be either a binary vector with frame-level labels (fully-supervised), a binary video-level label (weakly-supervised), a default one (one-class), or absent (unsupervised). 
However, in practice, it can be costly to collect $\classbinary$ as anomalies are rare, and $\video$ itself due to potential privacy concerns. Moreover, both label and video data may need regular updates due to evolving application contexts. 

Differently, in this paper, we introduce a novel setup for VAD, termed as \textit{training-free VAD}. Under this setting, we aim to estimate the anomaly score of each $\image \in \video$ using only pre-trained models at inference time, \ie without any training or fine-tuning involving a training dataset $\trainingset$.

\subsection{Are LLMs good for VAD?} 
\label{sec:llms}
\begin{figure*}[t!]
\centering
  \includegraphics[width=0.99\linewidth]{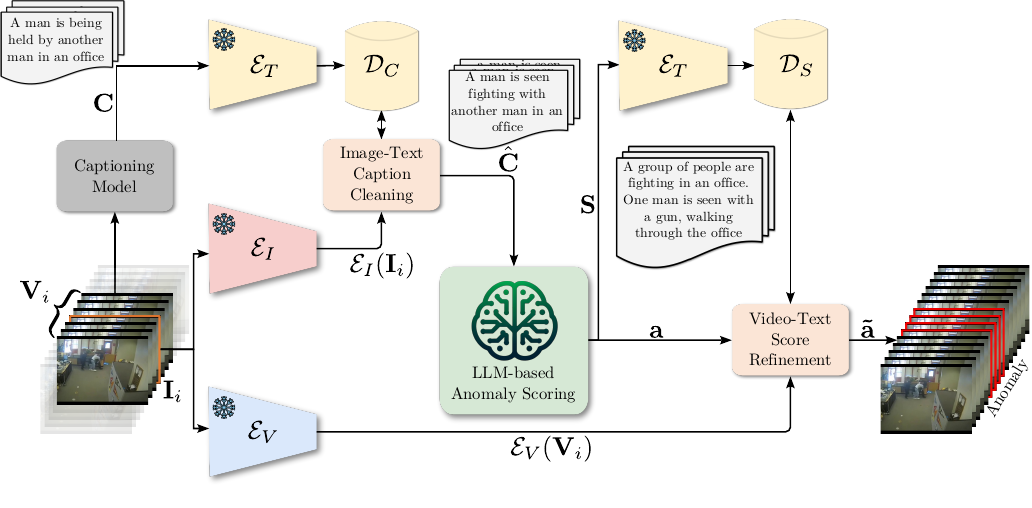}
  \vspace{-2mm}
  \caption{The architecture of our proposed~\methodshort~for addressing training-free VAD. For each test video $\video$, we first employ a captioning model to generate a caption $\rawcaption_i$ for each frame $\image_i \in \video$, forming a caption sequence $\rawcaptionset$. Our \textit{Image-Text Caption Cleaning} component addresses noisy and incorrect raw captions based on cross-modal similarity. We replace the raw caption with a caption $\retrievedcaption_i \in \rawcaptionset$ whose textual embedding $\netvlmtext(\retrievedcaption_i)$ is most aligned to the image embedding $\netvlmimage(\image_i)$, resulting in a cleaned caption sequence $\rawcaptionsetclean$. To account for scene context and dynamics, our \textit{LLM-based Anomaly Scoring} component further aggregates the cleaned captions within a temporal window centered around each $\image_i$ by prompting the LLM to produce a temporal summary $\summarycaption_i$, forming a summary sequence $\mathbf{\summarycaption}$. The LLM is then queried to provide an anomaly score for each frame based on its $\summarycaption_i$, obtaining the initial anomaly scores $\mathbf{\scoreabn}$ for all frames. Finally, our \textit{Video-Text Score Refinement} component refines each $\scoreabn_i$ by aggregating the initial anomaly scores of frames whose textual embeddings of the summaries are mostly aligned to the representation $\netvlmvideo(\video_i)$ of the video snippet $\video_i$ centered around $\image_i$, leading to the final anomaly scores $\mathbf{\refinedscoreabn}$ for detecting the anomalies (\inlineColorbox{AnomalyRed}{anomalous frames} are highlighted) within the video.}
 \label{fig:method}
\end{figure*}
We propose to address training-free VAD by exploiting recent advances in LLMs. As the use of LLMs in VAD is still in its infancy \cite{kim2023unsupervised}, we first analyze the capabilities of LLMs in producing an anomaly score based on a textual description of a video frame.

To achieve this, we first exploit a state-of-the-art captioning model $\netcaptioner$, \ie \bliptwo~\cite{li2023blip}, to generate a textual description for each frame $\image\in\video$. We then treat anomaly score estimation as a classification task, asking an LLM $\netllm$ to select only one score from a list of 11 uniformly sampled values in the interval $[0,1]$, where 0 means normal and 1 anomalous.
We get the anomaly score as:
\begin{equation}
    \label{eq:abnscore-preliminary}
\netllm(\prompt_C \circ \prompt_F \circ \netcaptioner(\image)) 
\end{equation}
where $\prompt_C$ is a context prompt that provides priors to the LLM regarding VAD, $\prompt_F$ instructs the LLM on the desired output format to facilitate automated text parsing\footnote{The exact form of $\prompt_F$ can be found in the \suppmat}, and $\circ$ is the text concatenation operation.
We devise $\prompt_C$ to simulate a potential end user of a VAD system, \eg law enforcement agency, as we empirically observe that impersonation can be an effective way of guiding the output generation of the LLM. 
For example, we can form $\prompt_C$ as: ``\textit{If you were a law enforcement agency, how would you rate the scene described on a scale from 0 to 1, with 0 representing a standard scene and 1 denoting a scene with suspicious activities?}".
Note that $\prompt_C$ does not encode any prior on the type of anomalies itself, but just on the context. 

Finally, with the estimated anomaly score from \cref{eq:abnscore-preliminary}, we measure the VAD performance using the standard area under the curve of the receiver operating characteristic (AUC ROC). \cref{fig:preliminary_bar} reports the results obtained on the test set of the UCF-Crime dataset~\cite{sultani2018real} with different variants of \bliptwo~ for obtaining the frame captions, and with different LLMs including \llama~\cite{touvron2023llama} and Mistral~\cite{jiang2023mistral} for computing the frame-level anomaly scores. For reference, we also provide the state-of-the-art performance under the unsupervised setting (the closest setting to ours) \cite{thakare2023dyannet}, and the random scoring as lower-bound. The plot demonstrates that state-of-the-art LLMs possess anomaly detection capabilities, largely outperforming random scoring.
However, this performance is much lower w.r.t. trained state-of-the-art methods, even in an unsupervised setting. 

We observe that two aspects might be the limiting factors in LLMs' performance. Firstly, the frame-level captions can be very noisy: the captions might be broken or may not fully reflect the visual content (see \cref{fig:preliminary_qualitative}). 
Despite the use of \bliptwo~\cite{li2023blip}, the best off-the-shelf captioning model, some captions appear corrupted, thus leading to unreliable anomaly scores. Secondly, the frame-level caption lacks details about the global context and the dynamics of the scene, which are key elements when modeling videos. 
In the following, we address these two limitations and propose \methodshort, the first training-free method for VAD that leverages LLMs for anomaly scoring together with modality-aligned VLMs.

%% file: sections/3_method.tex
\subsection{\methodshort: LAnguage-based VAD} 
\label{sec:method}

\methodshort~decomposes the VAD function $f$ into five elements (see \cref{fig:method}). As in the preliminary study, the first two are the captioning module $\netcaptioner$ mapping images to textual descriptions in the language space $\languagespace$, \ie $\netcaptioner:\imagespace\rightarrow\languagespace$, and the LLM $\netllm$ generating text from language queries, \ie $\netllm:\languagespace\rightarrow\languagespace$. 
The other elements involve three encoders mapping input representations to a shared latent space $\latentspace$. Specifically we have the image encoder $\netvlmimage:\imagespace\rightarrow\latentspace$, the textual encoder $\netvlmtext:\languagespace\rightarrow\latentspace$, and the video encoder
$\netvlmvideo:\videospace\rightarrow\latentspace$ for videos. 
Note that all five elements involve only off-the-shelf frozen models.
 
Following the positive findings of the preliminary analysis, \methodshort~leverages $\netllm$ and $\netcaptioner$ to estimate the anomaly score for each frame. We design \methodshort~to address the limitations related to noise and lack of scene dynamics in frame-level captions by introducing three components: i) Image-Text Caption Cleaning through the vision-language representations of $\netvlmimage$ and $\netvlmtext$, ii) LLM-based Anomaly Scoring, encoding temporal information via $\netllm$ and iii) Video-Text Score Refinement of the anomaly scores via video-text similarity, using $\netvlmvideo$ and $\netvlmtext$. In the following, we describe each component in detail.

\vspace{5pt}\noindent\textbf{Image-Text Caption Cleaning.} For each test video $\video$, we first employ $\netcaptioner$ to generate a caption $\rawcaption_i$ for each frame $\image_i \in \video$. Specifically, we denote as $\rawcaptionset=[\rawcaption_1,\ldots,\rawcaption_\numimages]$ the sequence of captions, where $\rawcaption_i = \netcaptioner(\image_i)$. 
However, as shown in \cref{sec:llms}, the raw captions can be noisy, with broken sentences or incorrect descriptions. 
To mitigate this issue, we rely on the captions of the whole video $\rawcaptionset$ assuming that in this set there exist captions that are unbroken and better capture the content of their respective frames, an assumption often verified in practice as the video features a scene captured by static cameras at a high frame rate. Thus, semantic content among frames can overlap regardless of their temporal distances. 
From this perspective, we treat caption cleaning as finding the \textit{semantically} closest caption to a target frame $\image_i$ within $\rawcaptionset$. 

Formally, we make use of vision-language encoders and form a set of caption embeddings by encoding each caption in $\rawcaptionset$ via $\netvlmtext$, \ie $\{\netvlmtext(\rawcaption_1),\ldots,\netvlmtext(\rawcaption_\numimages)\}$.
For each frame $\image_i \in \video$, we compute its closest semantic caption as:
\begin{equation}
\label{eq:caption-cleaning}
    \retrievedcaption_i = \arg\max_{\rawcaption \in \rawcaptionset} \langle \netvlmimage(\image_i) \cdot \netvlmtext(\rawcaption)\rangle,
\end{equation}
where $\langle\cdot,\cdot\rangle$ is the cosine similarity, and $\netvlmimage$ the image encoder of the VLM. 
We then build the cleaned set of captions as $\rawcaptionsetclean =[\retrievedcaption_1,\ldots, \retrievedcaption_\numimages]$, replacing each initial caption $\rawcaption_i$ with its counterpart $\retrievedcaption_i$ retrieved from $\rawcaptionset$. 
By performing the caption cleaning process, we can 
propagate the captions of frames that are semantically more aligned to the visual content, regardless of their temporal positioning, to improve or correct noisy descriptions. 

\vspace{5pt}
\noindent\textbf{LLM-based Anomaly Scoring.} \label{sec:method:summary}
The obtained caption sequence $\rawcaptionsetclean$, while being cleaner than the initial set, lacks temporal information. To overcome this, we leverage the LLM to provide temporal summaries. Specifically, we define a temporal window of $\clipduration$ seconds, centered around $\image_i$. 
Within this window, we uniformly sample $\framesperclip$ frames, forming a video snippet $\video_i$, and a caption sub-sequence $\rawcaptionsetclean_i =\{\retrievedcaption_n\}_{n=1}^{\framesperclip}$. We can then query the LLM with $\rawcaptionsetclean_i$ and a prompt $\prompt_S$ to get the temporal summary $\summarycaption_i$ centered on frame $\image_i$: 
\begin{equation}
\label{eq:summary}
\summarycaption_i = \netllm(\prompt_S \circ \rawcaptionsetclean_i) 
\end{equation} 
where the prompt $\prompt_S$ is formed as 
``\textit{Please summarize what happened in few sentences, based on the following temporal description of a scene. Do not include any unnecessary details or descriptions.}"\footnote{$\rawcaptionsetclean_i$ is represented as an ordered list, with items separated by \texttt{\textbackslash n}.}.

Coupling \cref{eq:summary} with the refinement process of \cref{eq:caption-cleaning}, we obtain a textual description of the frame ($\summarycaption_i$) which is semantically and temporally richer than $\rawcaption_i$. With $\summarycaption_i$, we can then query the LLM for estimating an anomaly score. 
Following the same prompting strategy described in \cref{sec:llms}, we ask $\netllm$ to assign to each temporal summary $S_i$ a score $\scoreabn_i$ in the interval $[0,1]$. 
We get the score as:
\begin{equation}
    \label{eq:abnscore}
\scoreabn_i = \netllm(\prompt_C \circ \prompt_F \circ S_i) 
\end{equation}
where, as in \cref{sec:llms},  $\prompt_C$ is a context prompt containing VAD contextual priors, and $\prompt_F$ provides information on the desired output format. 

\vspace{5pt}
\noindent\textbf{Video-Text Score Refinement.} %\label{sec:method:refinement}
By querying the LLM for each frame in the video with \cref{eq:abnscore}, we obtain the initial anomaly scores of the video $\mathbf{\scoreabn}=[\scoreabn_1,\ldots,\scoreabn_\numimages]$. However, $\mathbf{\scoreabn}$ is purely based on the language information encoded in their summaries, without taking into account the whole set of scores. Thus, we further refine them by leveraging the visual information to aggregate scores from semantically similar frames. 
Specifically, we encode the video snippet $\video_i$ centered around $\image_i$ using $\netvlmvideo$ and all the temporal summaries using $\netvlmtext$. 
Let us define $\mathbf{K}_i$ as the set of indices of the $\numk$-closest temporal summaries to $\video_i$ in $\{\summarycaption_1,\ldots,\summarycaption_\numimages\}$, where the similarity between $\video_i$ and a caption $\summarycaption_j$ is the cosine similarity, \ie $\langle\netvlmvideo(\video_i),\netvlmtext(\summarycaption_j)\rangle$. 
We obtain the refined anomaly score $\refinedscoreabn_i$:
%++++++++++++++++++++++++++++++++++++++++++++++++
\begin{equation}
    \label{eq:abnref}
    \refinedscoreabn_i = \sum_{k \in \mathbf{K}_i}{\scoreabn_k} \cdot \frac{e^{\langle\netvlmvideo(\video_i),\netvlmtext(\summarycaption_k)\rangle}}{\sum_{k \in \mathbf{K}_i}e^{\langle\netvlmvideo(\video_i),\netvlmtext(\summarycaption_k)\rangle}}
\end{equation}
%++++++++++++++++++++++++++++++++++++++++++++++++
where $\langle\cdot,\cdot\rangle$ is the cosine similarity. Note that \cref{eq:abnref} exploits the same principles of \cref{eq:caption-cleaning}, refining frame-level estimations (\ie score/captions) using their visual-language similarity (\ie video/image) with other frames in the video. 
Finally, with the refined anomaly scores for the test video $\mathbf{\refinedscoreabn}=[\refinedscoreabn_1,\ldots,\refinedscoreabn_\numimages]$, we identify the anomalous temporal windows via thresholding. 

%% file: sections/4_experiment.tex
\section{Experiments}\label{sec:exp}
%=======================================================
\begin{table}[t!]
\resizebox{0.95\linewidth}{!}{
\centering
\begin{tabular}{lcc}
\toprule
\textbf{\textsc{Method}} & \textbf{\textsc{Backbone}} & \textbf{\textsc{AUC(\%)}} \\
\midrule
\rowcolor{FunctionPurple} \textsc{Sultani \etal} \cite{sultani2018real} & C3D-RGB & 75.41\\
\rowcolor{FunctionPurple} \textsc{Sultani \etal} \cite{sultani2018real} & I3D-RGB & 77.92\\
\rowcolor{FunctionPurple} \textsc{IBL} \cite{zhang2019temporal} & \textsc{C3D-RGB} & 78.66\\
\rowcolor{FunctionPurple} \textsc{GCL} \cite{zaheer2022generative} & ResNext & 79.84\\
\rowcolor{FunctionPurple} \textsc{GCN} \cite{zhong2019graph} & TSN-RGB & 82.12\\
\rowcolor{FunctionPurple} \textsc{MIST} \cite{feng2021mist} & I3D-RGB & 82.30\\
\rowcolor{FunctionPurple} \textsc{Wu \etal}\cite{wu2020not} & I3D-RGB & 82.44\\
\rowcolor{FunctionPurple} \textsc{CLAWS} \cite{zaheer2020claws} & \textsc{C3D-RGB} & 83.03\\
\rowcolor{FunctionPurple}  \textsc{RTFM} \cite{tian2021weakly} & VideoSwin-RGB & 83.31\\
\rowcolor{FunctionPurple} \textsc{RTFM} \cite{tian2021weakly} & I3D-RGB & 84.03\\
\rowcolor{FunctionPurple} \textsc{Wu $\&$ Liu} \cite{wu2021learning} & I3D-RGB & 84.89\\
\rowcolor{FunctionPurple} \textsc{MSL} \cite{li2022self} & I3D-RGB & 85.30\\
\rowcolor{FunctionPurple} \textsc{MSL} \cite{li2022self} & VideoSwin-RGB & 85.62\\
\rowcolor{FunctionPurple} \textsc{S3R} \cite{wu2022self} & I3D-RGB & 85.99\\
\rowcolor{FunctionPurple} \textsc{MGFN} \cite{chen2023mgfn} & VideoSwin-RGB & 86.67\\
\rowcolor{FunctionPurple} \textsc{MGFN} \cite{chen2023mgfn} & I3D-RGB & 86.98\\
\rowcolor{FunctionPurple} \textsc{SSRL} \cite{li2022scale} & I3D-RGB & 87.43\\
\rowcolor{FunctionPurple} \textsc{CLIP-TSA} \cite{joo2023clip} & ViT & 87.58 \\
\midrule
\rowcolor{LanguageOrange}  \textsc{SVM} \cite{sultani2018real} & - & 50.00\\
\rowcolor{LanguageOrange}  \textsc{SSV} \cite{sohrab2018subspace} & - & 58.50\\
\rowcolor{LanguageOrange}  \textsc{BODS} \cite{wang2019gods} & I3D-RGB & 68.26\\
\rowcolor{LanguageOrange} \textsc{GODS} \cite{wang2019gods} & I3D-RGB & 70.46\\
\midrule
\rowcolor{VisionBlue} \textsc{GCL} \cite{zaheer2022generative} & ResNext & 74.20\\
\rowcolor{VisionBlue} \textsc{Tur \etal} \cite{tur2023exploring} & ResNet & 65.22\\
\rowcolor{VisionBlue} \textsc{Tur \etal} \cite{tur2023unsupervised} & ResNet & 66.85\\
\rowcolor{VisionBlue} \textsc{DyAnNet} \cite{thakare2023dyannet} & I3D & 79.76\\
\midrule
\rowcolor{ModelGreen} \textsc{ZS CLIP} \cite{radford2021learning} & ViT & 53.16\\
\rowcolor{ModelGreen} \textsc{ZS ImageBind (Image)} \cite{girdhar2023imagebind} & ViT & 53.65\\
\rowcolor{ModelGreen} \textsc{ZS ImageBind (Video)} \cite{girdhar2023imagebind} & ViT & 55.78\\
\rowcolor{ModelGreen} \textsc{LLaVA-1.5} \cite{liu2023improved} & ViT & 72.84\\
\rowcolor{ModelGreen}\textsc{\textbf{\methodshort}} &  ViT & \textbf{80.28}\\
\bottomrule
\end{tabular}
}
\caption{Comparison with state-of-the-art \inlineColorbox{FunctionPurple}{weakly-supervised}, \inlineColorbox{LanguageOrange}{one-class}, \inlineColorbox{VisionBlue}{unsupervised} and \inlineColorbox{ModelGreen}{training-free} methods on the UCF-Crime dataset. The best results among training-free methods are highlighted in bold.}
\label{table:ucf_results}
\end{table}
%=======================================================

%=======================================================
\begin{table}[t!]
\tabcolsep 3pt
\resizebox{\linewidth}{!}{
\centering
\begin{tabular}{lccc}
\toprule
 \textbf{\textsc{Method}} & \textbf{ \textsc{Backbone}} & \textbf{\textsc{AP(\%)}} & \textbf{\textsc{AUC(\%)}} \\
\midrule
\rowcolor{FunctionPurple} \textsc{Wu \etal} \cite{wu2020not} & C3D-RGB & 67.19 & -\\
\rowcolor{FunctionPurple} \textsc{Wu \etal} \cite{wu2020not} & I3D-RGB & 73.20 & -\\
\rowcolor{FunctionPurple} \textsc{MSL} \cite{li2022self} & C3D-RGB & 75.53 & -\\
\rowcolor{FunctionPurple} \textsc{Wu and Liu}\cite{wu2021learning} & I3D-RGB & 75.90 & -\\
\rowcolor{FunctionPurple} \textsc{RTFM} \cite{tian2021weakly} & I3D-RGB &77.81 & -\\
\rowcolor{FunctionPurple} \textsc{MSL} \cite{li2022self} & I3D-RGB & 78.28 & -\\
\rowcolor{FunctionPurple} \textsc{MSL} \cite{li2022self} & VideoSwin-RGB & 78.58 & -\\
\rowcolor{FunctionPurple} \textsc{S3R}\cite{wu2022self} & I3D-RGB & {80.26} & -\\
\rowcolor{FunctionPurple} \textsc{MGFN} \cite{chen2023mgfn} & I3D-RGB & 79.19 & -\\
\rowcolor{FunctionPurple} \textsc{MGFN} \cite{chen2023mgfn} & VideoSwin-RGB &80.11 & -\\
\midrule
\rowcolor{LanguageOrange} \textsc{Hasan \etal} \cite{hasan2016learning} & \text{AE\textsuperscript{RGB}} & - & 50.32\textsuperscript{$\ast$}\\
\rowcolor{LanguageOrange} \textsc{Lu \etal} \cite{lu2013abnormal} & Dictionary & - & 53.56\textsuperscript{$\ast$}\\
\rowcolor{LanguageOrange} \textsc{BODS} \cite{wang2019gods} & I3D-RGB & - & 57.32\textsuperscript{$\ast$}\\
\rowcolor{LanguageOrange} \textsc{GODS}\cite{wang2019gods} & I3D-RGB & - & 61.56\textsuperscript{$\ast$}\\
\midrule
\rowcolor{VisionBlue} \textsc{RareAnom} \cite{thakare2023rareanom} & I3D-RGB & - & 68.33\textsuperscript{$\ast$}\\
\midrule
\rowcolor{ModelGreen} \textsc{ZS CLIP} \cite{radford2021learning}  & ViT & 17.83 & 38.21\\
\rowcolor{ModelGreen} \textsc{ZS ImageBind (Image)} \cite{girdhar2023imagebind}  & ViT & 27.25 & 58.81\\
\rowcolor{ModelGreen} \textsc{ZS ImageBind (Video)} \cite{girdhar2023imagebind}  & ViT & 25.36 & 55.06\\
\rowcolor{ModelGreen} \textsc{LLaVA-1.5} \cite{liu2023improved} & ViT & 50.26 & 79.62\\
\rowcolor{ModelGreen} \textbf{\textsc{\methodshort}} &   ViT & \textbf{62.01} & \textbf{85.36}\\
\bottomrule
\end{tabular}
}
\caption{Comparison with state-of-the-art \inlineColorbox{FunctionPurple}{weakly-supervised}, \inlineColorbox{LanguageOrange}{one-class}, \inlineColorbox{VisionBlue}{unsupervised} and \inlineColorbox{ModelGreen}{training-free} methods on the XD-Violence dataset. $\ast$ denotes results reported in \cite{thakare2023rareanom}. The best results among training-free methods are highlighted in bold.}
\label{table:xd_violence_results}
\end{table}

We validate our training-free proposal \methodshort~on two datasets in comparison with state-of-the-art VAD methods that are trained with different levels of supervision, as well as training-free baselines.
We conduct an extensive ablation study to justify our main design choices regarding the proposed components, prompt design, and score refinement. 
In the following, we first describe our experimental setup in terms of datasets and performance metrics. We then present and discuss the results in \cref{sec:exp:comparison}, followed by the ablation study in \cref{sec:exp:ablation}. 
We show more qualitative results and ablation on minor designs in the \textit{Supp.~Mat.}

\vspace{2pt}
\noindent\textbf{Datasets.}
We evaluate our method using two commonly used VAD datasets featuring real-world surveillance scenarios, \ie UCF-Crime \cite{sultani2018real} and XD-Violence \cite{wu2020not}.

\noindent\textbf{\textit{UCF-Crime}} is a large-scale dataset that is composed of 1900 long untrimmed real-world surveillance videos, covering 13 real-world anomalies. The training set consists of 800 normal and 810 anomalous videos, while the test set includes 150 normal and 140 anomalous videos. 

\vspace{2pt}
\noindent\textbf{\textit{XD-Violence}} is another large-scale dataset for violence detection, comprising 4754 untrimmed videos with audio signals and weak labels that are collected from both movies and YouTube. XD-Violence captures 6 categories of anomalies and it is divided into a training set of 3954 videos and a test set of 800 videos.

\vspace{2pt}
\noindent\textbf{Performance Metrics.}
We measure the VAD performance using the area under the curve (AUC) of the frame-level receiver operating characteristics (ROC) as it is agnostic to thresholding for the detection task. For the XD-Violence dataset, we also report the average precision (AP), \ie the area under the frame-level precision-recall curve, following the established evaluation protocol in~\cite{wu2020not}.

\noindent\textbf{Implementation Details.} We sample each video every 16 frames for computational efficiency. We employ \bliptwo~\cite{li2023blip} as the captioning module $\netcaptioner$. Particularly, we consider an ensemble of \bliptwo~model variants in our Image-Text Caption Cleaning technique. Please refer to \suppmat~for a detailed analysis of these variants. We use Llama-2-13b-chat \cite{touvron2023llama} as our LLM module $\netllm$.
We use multimodal encoders provided by ImageBind~\cite{girdhar2023imagebind}. Specifically, the temporal window is $\clipduration=10$ seconds, in line with the pre-trained video encoder of ImageBind. 
We employ $\numk=10$ in Video-Text Score Refinement. 

\subsection{Comparison with state of the art}\label{sec:exp:comparison}
We compare \methodshort~against state-of-the-art approaches, including unsupervised methods~\cite{zaheer2022generative, tur2023exploring, tur2023unsupervised, thakare2023dyannet, thakare2023rareanom}, one-class methods~\cite{hasan2016learning,lu2013abnormal,wang2019gods,sultani2018real,sohrab2018subspace}, and weakly-supervised methods~\cite{sultani2018real,zhang2019temporal,zaheer2022generative,zhong2019graph,feng2021mist,wu2020not,zaheer2020claws,tian2021weakly,wu2021learning,li2022self,li2022self,wu2022self,chen2023mgfn,li2022scale,joo2023clip}. 
In addition, as none of the above methods specifically address VAD in a training-free setup, we further introduce a few training-free baselines with VLMs, \ie CLIP \cite{radford2021learning}, ImageBind \cite{girdhar2023imagebind}, and LLaVa \cite{liu2023improved}.

Specifically, we introduce Zero-shot CLIP \cite{radford2021learning} ($\textsc{ZS CLIP}$) and Zero-shot ImageBind \cite{girdhar2023imagebind} ($\textsc{ZS ImageBind}$). For both baselines, we exploit their pre-trained encoders to compute the cosine similarities of each frame embedding against the textual embeddings of two prompts: \textit{a standard scene} and \textit{a scene with suspicious or potentially criminal activities}. We then apply a softmax function to the cosine similarities to obtain the anomaly score for each frame. Since ImageBind also supports the video modality, we include $\textsc{ZS ImageBind (Video)}$ using the cosine similarities of the video embeddings against the two prompts. {We choose ViT-B/32 \cite{dosovitskiy2020vit} as the visual encoder for \textsc{ZS-CLIP}, ViT-H/14 \cite{dosovitskiy2020vit} as the visual encoders for \textsc{ZS-ImageBind (Image, Video)}, and both utilize CLIP's text encoder \cite{radford2021learning}.}
Finally, we introduce a baseline based on \textsc{LLaVA-1.5}, where we directly query LLaVa~\cite{liu2023improved} to generate an anomaly score for each frame, using the same context prompt as in ours. {\textsc{LLaVA-1.5} uses CLIP ViT-L/14 \cite{radford2021learning} as the visual encoder and Vicuna-13B as the LLM.}

\begin{figure*}[t!]
\centering
\includegraphics[width=\linewidth]{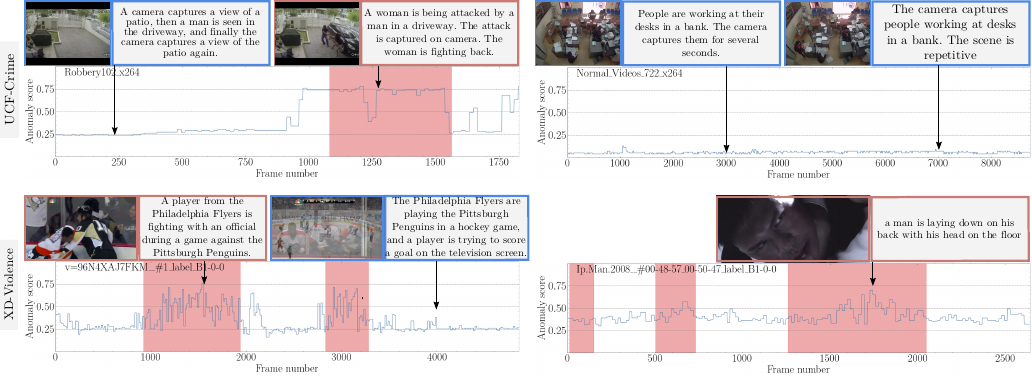}
  \caption{We showcase qualitative results obtained by \methodshort~on four test videos, including two videos (top row) from UCF-Crime and two videos from XD-Violence (bottom row). For each video, we plot the anomaly score over frames computed by our method. We display some keyframes alongside their most aligned temporal summary (blue bounding boxes for normal frame predictions and red bounding boxes for abnormal frame predictions), illustrating the relevance among the predicted anomaly score, visual content, and description. \inlineColorbox{AnomalyRed}{Ground-truth anomalies} are highlighted.}
 \label{fig:qualitatives}
\end{figure*}

\cref{table:ucf_results} presents the results of the full comparison against the state-of-the-art methods, as well as our introduced training-free baselines, on the UCF-Crime dataset \cite{sultani2018real}. 
Notably, our method without any training demonstrates superior performance compared to both the one-class and unsupervised baselines, achieving a higher AUC ROC, with a significant improvement of $+6.08\%$ when compared to GCL \cite{zaheer2022generative} and a minor improvement of $+0.52\%$ against the current state of the art obtained by DyAnNet~\cite{thakare2023dyannet}.

Moreover, it is evident that training-free VAD is a challenging task as a naive application of VLMs to VAD, such as \textsc{ZS CLIP}, \textsc{ZS ImageBind (Image)} and \textsc{ZS ImageBind (Video)}, leads to poor VAD performance. VLMs are mostly trained to attend to foreground objects, rather than actions or the background information in an image that contributes to the judgment of anomalies. This might be the main reason for the poor generalization of VLMs on the VAD task. 
The baseline \textsc{LLaVA-1.5}, which directly prompts for the anomaly score for each frame, achieves a much higher VAD performance than directly exploiting VLMs in a zero-shot manner. Yet, its performance is still inferior to ours, where we leverage a richer temporal scene description for anomaly estimation, instead of a single-frame basis. The similar effect of the temporal summary is also confirmed by our ablation study as presented in \cref{tab:ablation_components}.
We also report the comparison against state-of-the-art methods and our baselines evaluated on XD-Violence in \cref{table:xd_violence_results}. Ours achieves superior performance compared to all one-class and unsupervised methods. In particular, \methodshort~outperforms RareAnom~\cite{thakare2023rareanom}, the best-scoring unsupervised method, by a substantial margin of $+17.03\%$ in terms of AUC ROC.

\vspace{2pt}
\noindent\textbf{Qualitative Results.} 
\cref{fig:qualitatives} shows qualitative results of \methodshort~with sample videos from UCF-Crime and XD-Violence, where we highlight some keyframes with their temporal summaries.
In the three abnormal videos (Row 1, Column 1, and Row 2), we can see that the temporal summaries of the keyframes during the anomalies accurately portray the visual content regarding the anomalous situations, which in turn benefits \methodshort~to correctly identify the anomalies. In the case of \textit{Normal\_Videos\_722} (row 1, column 2), we can see that \methodshort~consistently predicts a low anomaly score throughout the video. For more qualitative results on the test videos, please refer to the \suppmat

%%%%%%%%%%%%%%%%%%%%%%%%%%%%%%%%%%%%%%%%%%%%
\subsection{Ablation study}\label{sec:exp:ablation}
In this section, we present the ablation study conducted with the UCF-Crime dataset. 
We first ablate the effectiveness of each proposed component of \methodshort. Then, we demonstrate the impact of task-related priors in the context prompt $\prompt_C$ when prompting the LLM for estimating the anomaly scores. Finally, we show the effect of $\numk$ when aggregating the $\numk$ semantically closest frames in the Video-Text Score Refinement component. 

\vspace{2pt}
\noindent\textbf{Effectiveness of each proposed component.} We ablate different variants of our proposed method \methodshort~to prove the effectiveness of the three proposed components, including Image-Text Caption Cleaning, LLM-based Anomaly Score, and Video-Text Score Refinement. \cref{tab:ablation_components} shows the results of all ablated variants of \methodshort. 
When the Image-Text Caption Cleaning component is omitted (Row 1), \ie the LLM only exploits the raw captions to perform temporal summary and obtain the anomaly scores with refinement, the VAD performance degrades by $-3.8 \%$ compared to \methodshort~ in terms of AUC ROC (Row 4). 
If we do not perform temporal summary, and only rely on the cleaned captions with refinement (Row 2), we observe a significant performance drop of $-7.58 \%$ compared to \methodshort~in AUC ROC, indicating that the temporal summary is an effective booster for LLM-based anomaly scoring. Finally, if we only use the anomaly scores obtained with the temporal summary on cleaned captions, without the final aggregation of semantically similar frames (Row 3), we can see that the AUC ROC decreases with a significant margin of $-7.49 \%$ compared to \methodshort, proving that Video-Text Score Refinement also plays an important role in improving the VAD performance. 

\vspace{2pt}
\noindent\textbf{Task priors in the context prompt.} We investigate the impact of different priors in the context prompt ${\prompt}_C$ and present the results in \cref{tab:ablation_prompt}. In particular, we experimented on two aspects, \ie impersonation and anomaly prior, which we believe can potentially benefit the estimation of LLM. Impersonation may help the LLM to process the input from the perspective of potential end users of a VAD system, while anomaly prior, \eg anomalies are criminal activities, may provide the LLM with a more relevant semantic context.
Specifically, we ablate \methodshort~with various context prompts ${\prompt}_C$. 
We begin with a base context prompt: ``\textit{How would you rate the scene described on a scale from 0 to 1, with 0 representing a standard scene and 1 denoting a scene with suspicious activities?}" (Row 1). 
We inject only the anomaly prior by appending ``\textit{suspicious activities}" with ``\textit{or potentially criminal activities}" (Row 2). We incorporate only impersonation by adding ``\textit{If you were a law enforcement agency,}'' at the beginning of the base prompt (Row 3).
Finally, we integrate both priors into the base context prompt (Row 4).
As shown in \cref{tab:ablation_prompt}, for videos within UCF-Crime, the anomaly prior appears to have a negligible effect on the LLM's assessment for anomalies, while impersonation improves the AUC ROC by $+0.96\%$ compared to the one obtained with only the base context prompt. Interestingly, incorporating both priors does not further boost the AUC ROC. We hypothesize that a more stringent context might limit the detection of a wider range of anomalies.

\begin{figure}[t!]
  \includegraphics[width=1.0\linewidth]{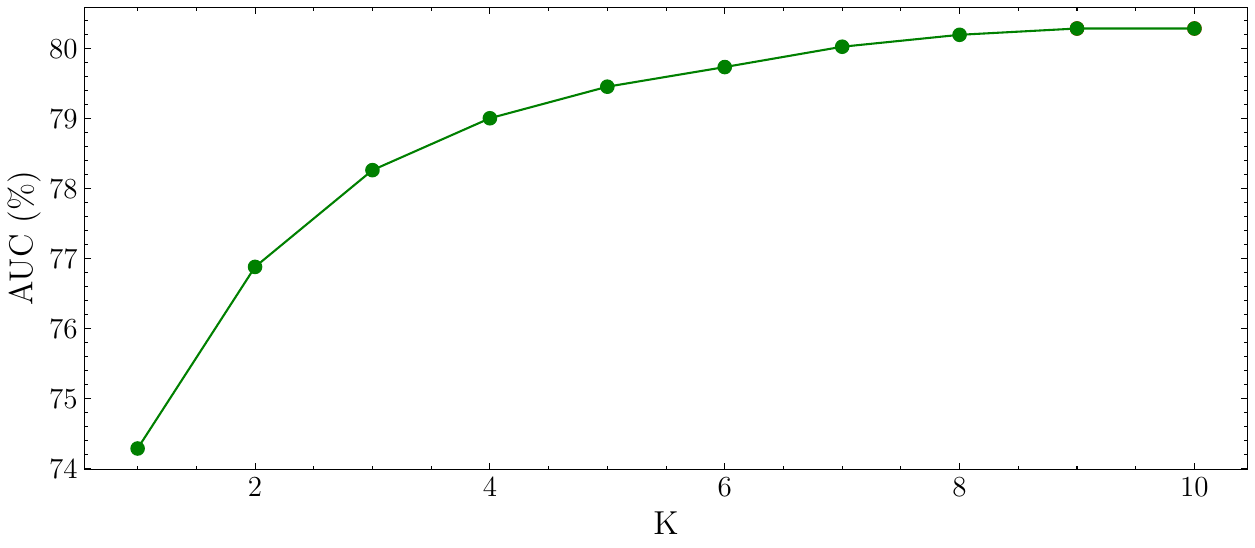}
  \caption{Results of \methodshort~on UCF-Crime over the number of $\numk$ semantically similar frames used for anomaly score refinement.}
 \label{fig:ablation_knn}
\end{figure}

\vspace{2pt}
\noindent\textbf{Effect of $\numk$ on refining anomaly score.} In this experiment, we investigate how the VAD performance changes in relation to the number of semantically similar temporal summaries, \ie $\numk$, used for refining the anomaly score of each frame. 
As depicted in \cref{fig:ablation_knn}, the AUC ROC metric consistently increases as $\numk$ increases, and saturates when $\numk$ approaches 9. The plot confirms the contribution of accounting semantically similar frames in obtaining more reliable anomaly scores of the video.

%=======================================================
\begin{table}[t!]
\centering
\resizebox{1\columnwidth}{!}{%
\begin{tabular}{cccc}
\toprule
\textbf{\textsc{Image-Text}} & \textbf{\textsc{LLM-based}}       & \textbf{\textsc{Video-Text}}        & \textbf{\textsc{AUC}}       \\ 
\textbf{\textsc{Caption Cleaning}} & \textbf{\textsc{Anomaly Scoring}}       & \textbf{\textsc{Score Refinement}}        & \textbf{\textsc{(\%)}}       \\ 
\midrule
     \xmark      & \cmark & \cmark & 76.48 \\
\cmark &     \xmark         & \cmark & 72.70\\
\cmark &  \cmark &   \xmark           & 72.79\\
\cmark & \cmark & \cmark & \textbf{80.28} \\
\bottomrule
\end{tabular}%
}
\caption{Results of \methodshort~variants w/o each proposed component on the UCF-Crime Dataset.}
\label{tab:ablation_components}
\end{table}
%=======================================================

%=======================================================
\begin{table}[t!]
\centering
\resizebox{0.8\columnwidth}{!}{%
\begin{tabular}{ccc}
\toprule
\textbf{\textsc{Anomaly prior}}   & \textbf{\textsc{Impersonation}} & \textbf{\textsc{AUC (\%)}}       \\ \midrule
\xmark & \xmark & 79.32\\
\cmark & \xmark & 79.38\\
\xmark & \cmark & \textbf{80.28} \\
\cmark & \cmark & 79.77 \\ 
\bottomrule
\end{tabular}%
}%
\caption{Results of \methodshort~on UCF-Crime with different priors in the context prompt when querying the LLM for anomaly scores.}
\label{tab:ablation_prompt}
\end{table}
%=======================================================

%% file: sections/5_conclusion.tex
\section{Conclusions}\label{sec:conclusion}
In this work, we introduced LAVAD, a pioneering method to address training-free VAD. \methodshort~follows a language-driven pathway for estimating the anomaly scores, leveraging off-the-shelf LLMs and VLMs. LAVAD has three main components, where the first uses image-text similarities to clean the noisy captions provided by a captioning model; the second leverages an LLM to aggregate scene dynamics over time and estimate anomaly scores; and the final component refines the latter by aggregating scores from semantically close frames according to video-text similarity. We evaluated \methodshort~on both UCF-Crime and XD-Violence, demonstrating superior performance compared to training-based methods in the unsupervised and one-class setting, without the need for training and additional data collection.

%% file: supp/supp.tex
In this supplementary material, we first provide the exact form of the prompts employed in our method and then we present additional experimental analyses.
Specifically, we first present the impact of the task-related priors in prompting the anomaly scores on XD-Violence \cite{wu2020not}. We then present the impact of captioning models, \ie different variants of \bliptwo~models, for the VAD performance of our method on both XD-Violence \cite{wu2020not} and UCF-Crime \cite{sultani2018real} datasets. Finally, we ablate the hyperparameters in constructing temporal windows to justify our design choice. Moreover, we describe the limitations and broader social impacts of our work, and we showcase additional qualitative results that demonstrate temporal summaries and the detection results. More qualitative results in the form of videos can be conveniently accessed on the project website at \url{https://lucazanella.github.io/lavad/}.

%% file: supp/prompts.tex
\section{Prompts}
\label{sec:prompts}

The prompts utilized in our approach serve distinct purposes. The contextual prompt $\prompt_C$ provides priors to the LLM for VAD. In line with the findings of our ablation studies presented in \cref{tab:ablation_prompt} and in \cref{tab:ablation_prompt_xd}, this prompt differs for UCF-Crime \cite{sultani2018real} and XD-Violence \cite{wu2020not}. For UCF-Crime, the prompt is structured as: ``\textit{If you were a law enforcement agency, how would you rate the scene described on a scale from 0 to 1, with 0 representing a standard scene and 1 denoting a scene with suspicious activities?}". In contrast, for XD-Violence, the prompt has the form: ``\textit{How would you rate the scene described on a scale from 0 to 1, with 0 representing a standard scene and 1 denoting a scene with suspicious or potentially criminal activities?}".

The prompt $\prompt_F$ provides guidance to the LLM for the desired output format, aimed at facilitating automated text parsing. This prompt remains consistent across both datasets and is defined as follows: ``\textit{Please provide the response in the form of a Python list and respond with only one number in the provided list below [0, 0.1, 0.2, 0.3, 0.4, 0.5, 0.6, 0.7, 0.8, 0.9, 1.0] without any textual explanation. It should begin with `[' and end with `]'.}".

Lastly, the prompt $\prompt_S$ is employed to obtain a temporal summary $\summarycaption_i$ for each frame $\image_i$. The prompt is formulated as follows: ``\textit{Please summarize what happened in few sentences, based on the following temporal description of a scene. Do not include any unnecessary details or descriptions.}".

%% file: supp/ablation.tex
\section{Additional analyses}
\label{sec:supp_ablation}

\noindent\textbf{Task priors in the context prompt.}
In \cref{tab:ablation_prompt_xd} we present the impact of different priors in the context prompt ${\prompt}_C$, i.e. impersonation and anomaly priors, on XD-Violence \cite{wu2020not}. This follows the same ablation design as presented in \cref{tab:ablation_prompt} in the main manuscript for UCF-Crime, with the priors added in the same way for both datasets.
As shown in \cref{tab:ablation_prompt_xd}, for videos within XD-Violence, incorporating the anomaly prior (Row 2) improves the average precision (AP) by $+1.67\%$ compared to using only the base context prompt (Row 1). Conversely, introducing impersonation (Row 3) degrades the AP by $-1.51\%$ compared to not using it (Row 1). Videos in XD-Violence originate from various sources, including CCTV cameras, movies, sports, and games. 
The effectiveness of the impersonation prior might be limited to CCTV camera videos, given that the surveillance domain is more closely associated with the concept of \textit{``law enforcement agency''} which is utilized for impersonation. Finally, combining both priors (Row 4) leads to improved performance compared to not utilizing any of them, primarily due to the positive impact of the anomaly prior.

%=======================================================
\begin{table}[t!]
\centering
\caption{Results of \methodshort~on XD-Violence with different priors in the context prompt when querying the LLM for anomaly scores.}
\label{tab:ablation_prompt_xd}
\resizebox{\columnwidth}{!}{%
\begin{tabular}{cccc}
\toprule
\textbf{\textsc{Anomaly prior}} & \textbf{\textsc{Impersonation}} & \textbf{\textsc{AP (\%)}} & \textbf{\textsc{AUC (\%)}} \\ \midrule
\xmark & \xmark & 60.34          & 84.42          \\
\cmark & \xmark & \textbf{62.01} & \textbf{85.36} \\
\xmark & \cmark & 58.83          & 84.50          \\
\cmark & \cmark & 60.78          & 85.26          \\ \bottomrule
\end{tabular}%
}
\end{table}
%=======================================================

\vspace{0.1cm}
\noindent\textbf{Impact of different \bliptwo~models.} 
As captioners, we consider different \bliptwo~\cite{li2023blip} models and their ensemble for both UCF-Crime \cite{sultani2018real} and XD-Violence \cite{wu2020not}, and we present the results in \Cref{tab:ablation_blip2_ucf_crime,tab:ablation_blip2_xd_violence}, respectively.

In \cref{tab:ablation_blip2_ucf_crime}, the most effective strategy for UCF-Crime videos is employing an ensemble of all \bliptwo~models (Row 6). This involves generating captions for all frames in a video using all \bliptwo~models and relying on the vision-language model (VLM) to identify the semantically closest captions for each frame. The effectiveness of the ensemble might be attributed to the challenges posed by UCF-Crime videos. These videos, characterized by low-resolution CCTV footage, often lead captioning models to hallucinate scene descriptions. For instance, it is common to encounter captions, such as \textit{``a person riding a skateboard down a road''} when the image only depicts a road in the absence of any specific event. The ensemble approach, by allowing the selection from a larger set of candidates, increases the likelihood of choosing more correct captions and filtering incorrect ones.

For XD-Violence, as shown in \cref{tab:ablation_blip2_xd_violence}, utilizing the captions generated by \textit{flan-t5-xxl} (Row 3) yields the best average precision (AP). 
Other \bliptwo~variants for XD-Violence may provide captions that prioritize foreground objects, potentially overlooking background elements constituting anomalies (\eg a vehicle enveloped in smoke on a busy street), yet better aligning with the VLM's representation of the video frames. Hence, when employing the ensemble of \bliptwo~models (Row 6), captions that specifically highlight elements constituting anomalies are not chosen as the semantically closest captions to video frames in the cleaning step, with a negative impact on the anomaly scoring phase. 

%=======================================================
\begin{table}[t!]
\centering
\caption{Results of \methodshort~on UCF-Crime with different \bliptwo~model variants in our Image-Text Caption Cleaning technique.}
\label{tab:ablation_blip2_ucf_crime}
\resizebox{\columnwidth}{!}{%
\begin{tabular}{cccccc}
\toprule
\multicolumn{5}{c}{\textbf{\textsc{\bliptwo}}} & \textbf{\textsc{AUC}} \\
\textbf{\textsc{flan-t5-xl}} &
  \textbf{\textsc{flan-t5-xl-coco}} &
  \textbf{\textsc{flan-t5-xxl}} &
  \multicolumn{1}{l}{\textbf{\textsc{opt-6.7b}}} &
  \multicolumn{1}{l}{\textbf{\textsc{opt-6.7b-coco}}} &
  \textbf{\textsc{(\%)}} \\ \midrule
\cmark  & \xmark  & \xmark & \xmark & \xmark & 74.19                 \\
\xmark  & \cmark  & \xmark & \xmark & \xmark & 74.49                 \\
\xmark  & \xmark  & \cmark & \xmark & \xmark & 74.38                 \\
\xmark  & \xmark  & \xmark & \cmark & \xmark & 75.50                 \\
\xmark  & \xmark  & \xmark & \xmark & \cmark & 73.94                 \\
\cmark  & \cmark  & \cmark & \cmark & \cmark & \textbf{80.28}                 \\ \bottomrule
\end{tabular}%
}
\end{table}
%=======================================================

%=======================================================
\begin{table}[t!]
\centering
\caption{Results of \methodshort~on XD-Violence with different \bliptwo~model variants in our Image-Text Caption Cleaning technique.}
\label{tab:ablation_blip2_xd_violence}
\resizebox{\columnwidth}{!}{%
\begin{tabular}{ccccccc}
\toprule
\multicolumn{5}{c}{\textbf{\textsc{\bliptwo}}} & \textbf{\textsc{AP}} & \textbf{\textsc{AUC}} \\
\textbf{\textsc{flan-t5-xl}} &
  \textbf{\textsc{flan-t5-xl-coco}} &
  \textbf{\textsc{flan-t5-xxl}} &
  \multicolumn{1}{l}{\textbf{\textsc{opt-6.7b}}} &
  \multicolumn{1}{l}{\textbf{\textsc{opt-6.7b-coco}}} &
  \textbf{\textsc{(\%)}} &
  \textbf{\textsc{(\%)}} \\ \midrule
\cmark  & \xmark  & \xmark & \xmark & \xmark & 61.09                & 85.16                 \\
\xmark  & \cmark  & \xmark & \xmark & \xmark & 57.41                & 82.78                 \\
\xmark  & \xmark  & \cmark & \xmark & \xmark & \textbf{62.01}                & \textbf{85.36}                 \\
\xmark  & \xmark  & \xmark & \cmark & \xmark & 56.55                & 82.42                 \\
\xmark  & \xmark  & \xmark & \xmark & \cmark & 54.71                & 82.93                 \\
\cmark  & \cmark  & \cmark & \cmark & \cmark & 59.62                & 84.90                 \\ \bottomrule
\end{tabular}%
}
\end{table}
%=======================================================

%=======================================================

%=======================================================
\vspace{0.1cm}
\noindent\textbf{Temporal window's duration and number of sampled frames.} In \cref{tab:ablation_T_and_N}, we evaluate the impact of varying the duration of the temporal window ($\clipduration$) and the number of sampled frames ($\framesperclip$), which is used to query the LLM for the temporal summary $\summarycaption_i$. Specifically, the temporal window duration $\clipduration$ determines the time interval, while the number of sampled frames $\framesperclip$ determines the number of captions. First, we conduct experiments by adjusting the duration $\clipduration$ to 2.5, 5, 10, and 20 seconds, while maintaining $\framesperclip = 10$. 
The 10-second temporal window yields the highest AUC score (Row 3). This is in line with the fact that ImageBind \cite{girdhar2023imagebind} is trained with video clips of 10 seconds.

Subsequently, we maintain the temporal window's duration $\clipduration$ at 10 seconds and vary the number of frames from 5 to 10 and 20. Notably, using 10 frames (Row 3), \ie 1 frame every second, is the optimal choice within this experiment. 
Balancing the number of captions per snippet presents a trade-off with the quality of the summary. Too many captions may overwhelm with excessive and non-diverse content, while too few captions may result in limited coverage of the content.

\begin{table}[t!]
\centering
\caption{Results of \methodshort~on UCF-Crime with different combinations of temporal window duration ($\clipduration$) and number of sampled frames per window ($\framesperclip$).}
\label{tab:ablation_T_and_N}
\resizebox{0.4\columnwidth}{!}{%
\begin{tabular}{ccc}
\toprule
\textbf{\textsc{\clipduration (s)}} & \textbf{\textsc{\framesperclip}} & \textbf{\textsc{AUC (\%)}} \\ \midrule
2.5 & 10 & 79.33          \\
5   & 10 & 78.10          \\
10  & 10 & \textbf{80.28} \\
20  & 10 & 79.24          \\ \midrule
10  & 5  & 77.48          \\
10  & 20 & 74.45          \\ \bottomrule
\end{tabular}%
}
\end{table}
%=======================================================

%% file: supp/qualitatives.tex
\section{Qualitative results}
\label{sec:qualitatives}

In \cref{fig:supp_qualitatives}, we present additional qualitative results demonstrating the effectiveness of our proposed \methodshort~in detecting anomalies within a set of UCF-Crime \cite{sultani2018real} and XD-Violence \cite{wu2020not} test videos. The figure showcases keyframes along with the most semantically similar temporal summaries. 
For example, in the video \textit{Shooting047} (Row 1, Column 1), \methodshort~assigns a high anomaly score when the video is labeled abnormal. However, it also assigns a high anomaly score during the initial and final segments, despite these parts being labeled as normal. This discrepancy arises because the video begins with text describing the subsequent content, leading the LLM to attribute a high anomaly score. In the final part, our method correctly identifies abnormality as the frame depicts a person on the ground who has been shot. In the video \textit{Burglary079} (Row 1, Column 2), there is a false abnormal instance. This occurs because the temporal summary associated with that frame incorrectly suggests the presence of a man stealing a car. In reality, the video depicts a man behaving suspiciously near the car, leading to a wrong interpretation by the captioning module. In the XD-Violence videos (Row 2), an anomaly caused by an explosion is correctly detected (Row 2, Column 1), while a normal video is consistently predicted as normal for more than $17,500$ frames (Row 2, Column 2). 

%% file: supp/limitations.tex
\section{Limitations}
\label{sec:limitations}

We identify two main limitations of our work. Firstly, our method fully relies on pre-trained models from VLMs and LLMs, thus its performance greatly depends on i) how well the captioning model describes the visual content, ii) how reliable the LLM is when generating the anomaly scores, and iii) how aligned the multi-modal encoders are when processing videos from various domains. Secondly, our anomaly scores are primarily obtained via prompting LLMs. Although we conducted experiments investigating different prompting strategies, a systematic understanding of LLM prompting for VAD requires a community effort.  

%% file: supp/broader_societal_impacts.tex
\section{Broader Societal Impacts}
\label{sec:broader_societal_impacts}

While our work pioneers the technical aspect of leveraging LLMs for detecting anomalies in videos, there exist open ethical challenges for a broader concern. VAD systems are mostly applied to safety-related contexts, for private use or public interests. Prior to any deployment, it is crucial to first investigate the behaviors of LLM-based methods, mitigating any potential bias in LLMs and improving explainability. {Our work serves as the first technical exploration of leveraging LLMs for training-free VAD, proving it as a competitive alternative. This is a necessary step to increase the awareness of the community on these important topics.}